\documentclass[runningheads]{llncs}
\usepackage[T1]{fontenc}
\usepackage{booktabs}
\usepackage{graphicx}
\usepackage[accsupp]{axessibility}  % Improves PDF readability for those with disabilities.

\usepackage{pifont}    % For checkmark and cross symbols
\newcommand{\cmark}{\ding{51}}  % Checkmark symbol
\newcommand{\xmark}{\ding{55}} 

\usepackage[pagebackref,breaklinks,colorlinks]{hyperref}
\usepackage{orcidlink}
\usepackage{subcaption}

\begin{document}
\title{R3ST: A Synthetic 3D Dataset With Realistic Trajectories}

\author{Simone Teglia\inst{1} \and Claudia Melis Tonti\inst{1} \and Francesco Pro\inst{1} \and Leonardo Russo\inst{1} \and Andrea Alfarano\inst{2} \and Matteo Pentassuglia\inst{3} \and Irene Amerini\inst{1} }
\authorrunning{Teglia et al.}
\institute{Sapienza University of Rome \and
INSAIT, Sofia University \and EURECOM, Biot, France \\ \texttt{\{teglia, melistonti, pro, russo, amerini\}@diag.uniroma1.it}, \texttt{andrea.alfarano@insait.ai}, \\ 
\texttt{russo.2015563@studenti.uniroma1.it}, \texttt{matteo.pentassuglia@eurecom.fr}
}

\maketitle              % typeset the header of the contribution
\begin{abstract}
Datasets are essential to train and evaluate computer vision models used for traffic analysis and to enhance road safety. Existing real datasets fit real-world scenarios, capturing authentic road object behaviors, however, they typically lack precise ground-truth annotations. In contrast, synthetic datasets play a crucial role, allowing for the annotation of a large number of frames without additional costs or extra time. However, a general drawback of synthetic datasets is the lack of realistic vehicle motion, since trajectories are generated using AI models or rule-based systems. In this work, we introduce \textbf{R3ST} (Realistic 3D Synthetic Trajectories), a synthetic dataset that overcomes this limitation by generating a synthetic 3D environment and integrating real-world trajectories derived from SinD, a bird's-eye-view dataset recorded from drone footage.
The proposed dataset closes the gap between synthetic data and realistic trajectories, advancing the research in trajectory forecasting of road vehicles, offering both accurate multimodal ground-truth annotations and authentic human-driven vehicle trajectories. We publicly release our dataset here. \footnote{\url{https://r3st-website.vercel.app/}} 

\end{abstract}

\section{Introduction}
\label{sec:intro}

Understanding and studying complex urban traffic scenarios is vital for traffic analysis, urban planning, and enhancing road safety. In particular, urban intersections poses significant challenges for Traffic Monitoring Systems (TMS) due to their complex layouts and dynamic traffic conditions, as they are the most accident-prone locations in urban environments. A significant hurdle in achieving this safety is the need for a comprehensive understanding of complex traffic environments and the precise prediction of other traffic participants movements. The advancement of predicting the motion of road agents is therefore vital for developing advanced traffic monitoring systems.
The analysis of traffic dynamics requires extensive datasets that effectively integrate the behavior of actual traffic participants with comprehensive annotations. Previous works like \cite{katariya2023pov} \cite{krajewski2018highd}\cite{9827305}\cite{sun2020scalability} either failed to address the inherently sophisticated nature of intersections, proposing datasets that contain real trajectories but lacking the necessary realism of an intersection scenario, or presented HD-Maps based datasets, that are not always suitable for intelligent TMS due to the absence of such precise data in real time.

\begin{figure}[htbp]
  \centering
  \begin{subfigure}[b]{0.49\textwidth}
    \includegraphics[alt={Simulated urban intersection with a bus, car, van, and motorcycles navigating through a multi-lane crossing.},width=\linewidth]{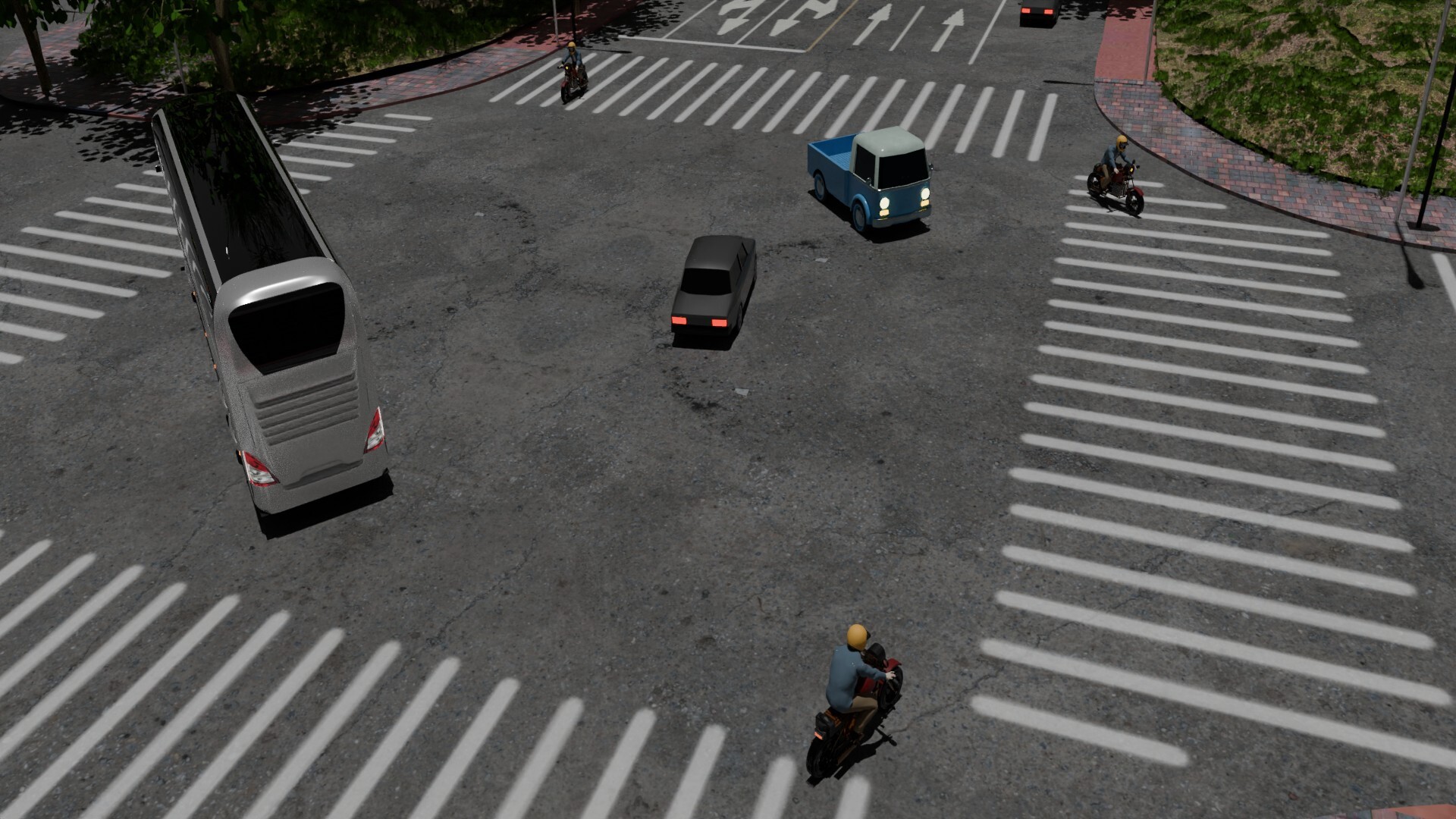}
  \end{subfigure}
  \hfill
  \begin{subfigure}[b]{0.49\textwidth}
    \includegraphics[alt={Depth map of an urban scene showing vehicles and cyclists, with closer objects appearing lighter and distant objects darker},width=\linewidth]{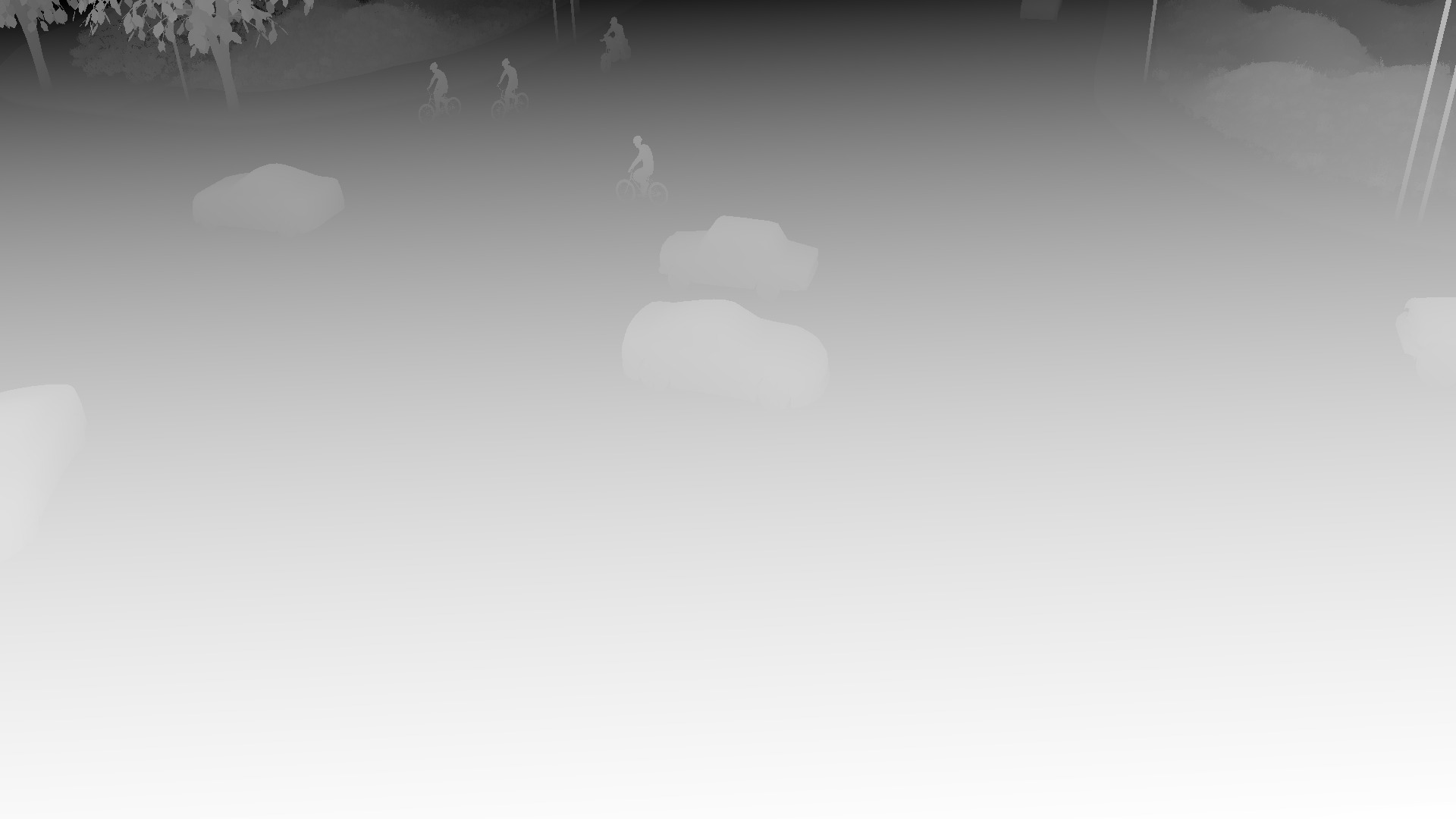}
  \end{subfigure}

  \vspace{0.2cm}

  \begin{subfigure}[b]{0.49\textwidth}
    \includegraphics[alt={Segmentatio map of an urban scene showing three cars crossing an intersection. The cars have the same color in the segmentation map.}, width=\linewidth]{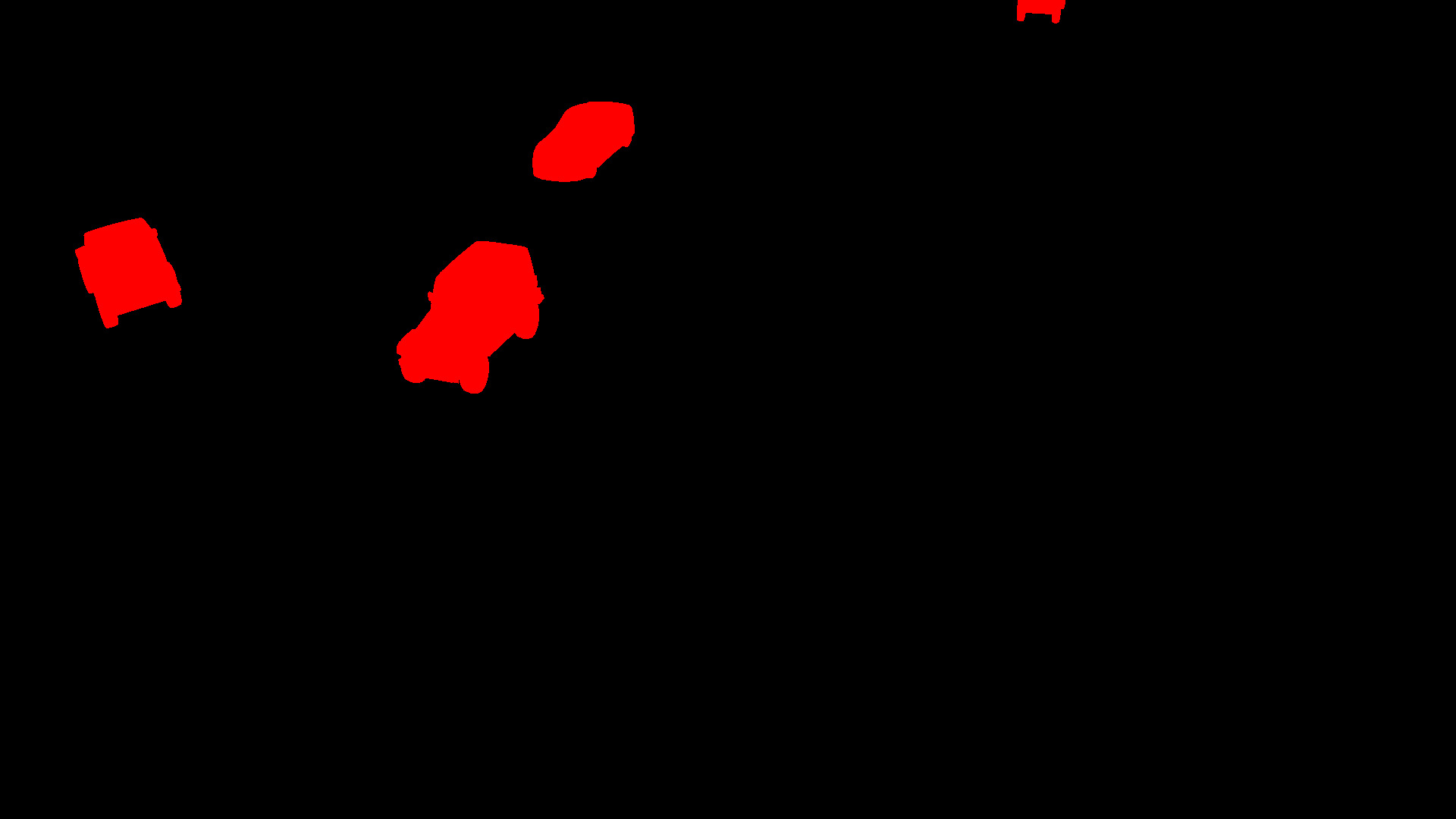}
  \end{subfigure}
  \hfill
  \begin{subfigure}[b]{0.49\textwidth}
    \includegraphics[alt={Simulated urban intersection with cars and bicycles. Around each vehicle is drawn a bounding box used for object detection.}, width=\linewidth]{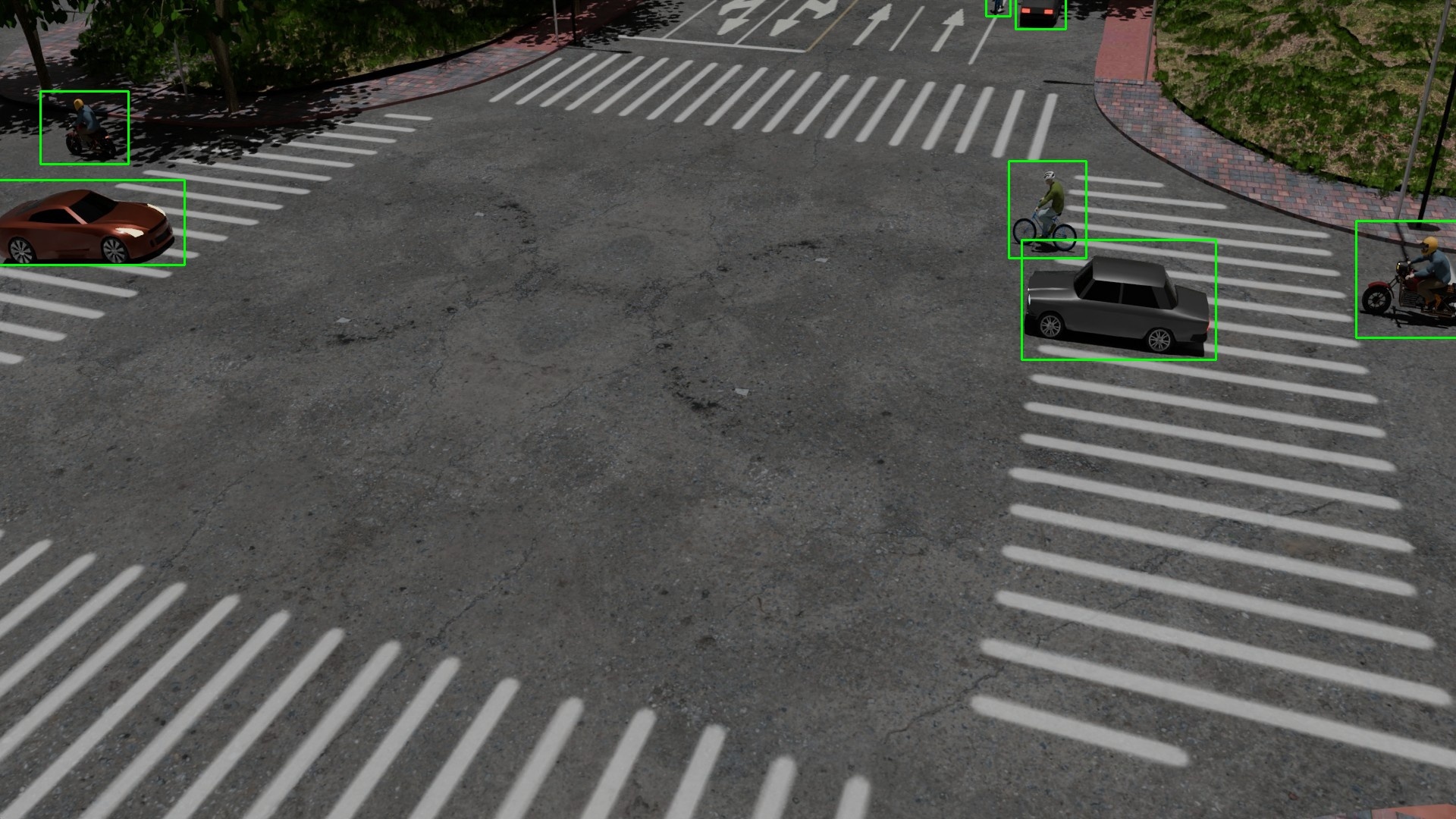}
  \end{subfigure}
  \caption{The R3ST Dataset. It provides photorealistic synthetic images (first), depth maps (second), instance segmentation masks (third), and object detection bounding boxes (fourth), enabling diverse computer vision tasks. Each task is presented on a different frame.}
  \label{r3st_images}

\end{figure}

The scarcity of a comprehensive dataset reflecting real road agent behavior poses serious challenges in creating precise and universally applicable motion prediction models, thereby constraining the efficacy of intelligent traffic monitoring in practical situations.
The behavior of drivers can noticeably deviate from the expected patterns due to the intrinsic unpredictability of their intentions, bad habits, poor road design, or high traffic levels. Those factors make simulation-based traffic models insufficient to completely capture the full spectrum of human responses to varying driving conditions, despite the considerable advances of simulation environments.
To address these limitations and bridge the gap between realism and precise ground-truth annotations, we introduce R3ST (Realistic 3D Synthetic Trajectories), a novel large-scale synthetic dataset that uniquely integrates real, human-driven trajectories derived from actual aerial drone footage (the SinD dataset \cite{xu2022drone}) into a constructed digital twin simulation. By leveraging real-world vehicle trajectories, R3ST captures authentic traffic behaviors while providing comprehensive multimodal annotations, such as instance segmentation, depth maps, and bounding boxes, facilitated by photorealistic rendering techniques and advanced annotation tools like Vision Blender \cite{cartucho2020visionblender}.

Specifically, R3ST features diverse urban intersections modeled and rendered within realistic environments created with Blender \cite{blender}. Unlike conventional synthetic datasets relying solely on rule-based or AI-generated vehicle motion, R3ST's vehicles follow real-world annotated trajectories, reflecting genuine human decision-making dynamics. 
%Unlike conventional synthetic datasets that rely solely on rule-based or AI-generated vehicles motion, R3ST's vehicles precisely follow real-world annotated trajectories, reflecting genuine human decision-making dynamics.
Furthermore, multiple camera angles, realistic sensor parameters, and high-quality rendering settings ensure that R3ST closely mimics actual road-camera images, significantly enhancing the dataset's applicability and value for diverse computer vision tasks.
Additionally, we leverage R3ST to evaluate pre-trained deep-learning models focusing on critical tasks like vehicle detection, instance segmentation, and monocular depth estimation. 
Through the innovative fusion of realistic human-driven trajectories and precise synthetic data generation, R3ST represents an advancement over previous datasets, offering researchers optimal capabilities for evaluating and training autonomous driving models, trajectory forecasting algorithms, and traffic analysis tools. The dataset presented in this paper offers critical foundations for future research in Intelligent Transportation Systems and contribute directly to safer, smarter urban mobility.

\begin{figure}[!h]
    \centering
    \includegraphics[alt={Trajectories of vehicles at a four-way intersection, visualized as colored paths representing different movement patterns and directions.}, width=0.8\linewidth]{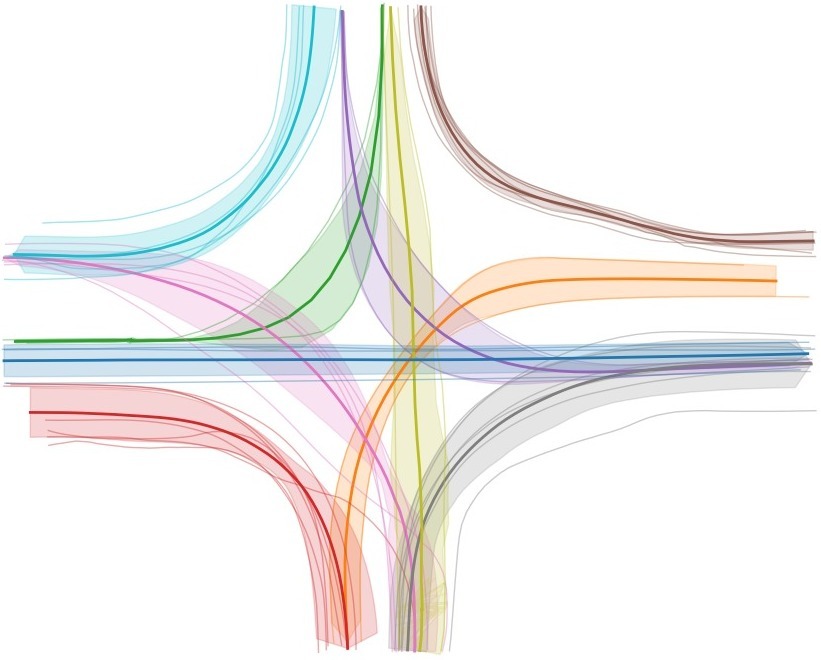}
    \caption{Visualization of clustered vehicle trajectories in a R3ST crossroad. Each colored trajectory represents a cluster of similar paths, while the shaded regions indicate variance within each cluster.}
    \label{fig:trajectories}
\end{figure}

\section{Related Work}

Datasets are essential for developing and optimizing Intelligent Traffic Monitoring Systems, particularly in complex urban scenarios such as intersections. These environments pose unique challenges due to the unpredictable behavior of road users, varying lighting conditions, and the need for real-time decision-making. To address these problems, a variety of datasets have been proposed, ranging from real-world to synthetic data. In this section, we review existing datasets used for road agents' perception and motion prediction, highlighting their strengths and weaknesses. To the best of our knowledge, no existing work has explored synthetic datasets that integrate real trajectories.

\paragraph{Real-World Datasets}
Real-world datasets are collected from sensors like cameras, LiDAR, or GPS, that are placed in real-world traffic situations.
Earlier works in the field of urban scene understanding like KITTI \cite{Geiger2013IJRR}, CityScapes \cite{Cordts2016Cityscapes}, SemanticKITTI \cite{behley2019iccv}, nuScenes \cite{vora2020pointpaintingsequentialfusion3d}, the Waymo Open Dataset \cite{sun2020scalability} and Argoverse 2 \cite{wilson2023argoverse2generationdatasets} provided complex scenarios allowing tasks like object detection, semantic segmentation and trajectory prediction, but from an on-board Point Of View (POV) or relying on HD-Maps. UAVDT \cite{du2018unmanned}, highD \cite{krajewski2018highd}, DroneVehicle \cite{sun2022drone}, and SinD \cite{xu2022drone} instead contain images recorded by drones, giving a new point of view with respect to previous datasets, addressing object detection and object tracking. 
Notable effort have been made by TUMTraf \cite{tumtraf}, which contains 4.8k humanly annotated frames captured by two LiDAR and two cameras pointed at an intersection.
However, the creation of a real-world dataset with precise annotations of vehicle trajectories remains an ongoing challenge, due to the difficulties and the high costs of acquiring large-scale, high-quality trajectory data.

\paragraph{Synthetic Datasets}
The problem of collecting real-world images and annotating them correctly, with an high expense of cost and time, has led to the creation of synthetic datasets.
Synthetic datasets have the advantage of being fully controllable. They allow the generation of diverse traffic scenarios and perfectly annotate frames since the attributes of each object in the scene, like position, size, and movement, are known to the generating environment. This, without additional costs or extra time, ensures consistency and eliminates human errors in the annotation process.
VIPER \cite{richter2017playingbenchmarks} and SYNTHIA \cite{ros2016synthia} exploited game engines like Grand Theft Auto V or Unity to generate synthetic datasets.
IDDA \cite{alberti2020idda}, SHIFT \cite{sun2022shift}, AIODrive \cite{Weng2020_AIODrive} and TUMTraf Synthetic \cite{tumtrafsynthetic} instead rely on CARLA \cite{dosovitskiy2017carlaopenurbandriving}, a popular simulation environment built on top of Unreal Engine 5, to create large-scale synthetic datasets for a variety of tasks, addressing the problem of the lack of various domains in a single dataset. Similarly, SynTraC \cite{chen2024syntrac} and Omni-IMOT \cite{sun2020simultaneous} are among the CARLA-based datasets the most closely relate to our work, as they focus specifically on intersections. 
% However, like other CARLA-based datasets, they still rely on simulated agent behaviors, which are inherently rule-based and lack the variability and unpredictability of real-world driving patterns. 
However, a significant limitation of synthetic datasets lies in the realism of vehicle movements. CARLA or game engines fail to recreate realistic trajectories since their movement relies on rule-based approaches that cannot model the unpredictability of human decision-making. This discrepancy poses important challenges in developing Intelligent TMS that are only trained on synthetic data, as models may struggle to generalize to real-world traffic scenarios. 

\begin{table*}[!ht]
    \centering
    \resizebox{\textwidth}{!}{  % Resize table to fit within the text width
    \begin{tabular}{l l c c c c c c}
        \toprule
        \textbf{Dataset} & \textbf{Year} & \textbf{Size} & \textbf{Real/Synthetic} & \textbf{POV} & \textbf{Depth} & \textbf{Instance Segmentation} & \textbf{Realistic Trajectories} \\
        \midrule
        KITTI \cite{Geiger2013IJRR} & 2012 & 41K frames & Real & On-Board & \textcolor{red}{\xmark} & \textcolor{red}{\xmark} & \textcolor{red}{\xmark} \\
        nuScenes \cite{vora2020pointpaintingsequentialfusion3d} & 2019 & 40K frames & Real & On-Board & \textcolor{red}{\xmark} & \textcolor{red}{\xmark} & \textcolor{red}{\xmark} \\
        SinD \cite{xu2022drone} & 2019 & 7 hours & Real & Drone & \textcolor{red}{\xmark} & \textcolor{red}{\xmark} & \textcolor{green}{\cmark} \\
        Argoverse2 \cite{wilson2023argoverse2generationdatasets} & 2023 & 2M frames & Real & On-Board & \textcolor{green}{\cmark} & \textcolor{red}{\xmark} & \textcolor{green}{\cmark} \\
        TUMTraf-I \cite{tumtraf} & 2023 & 4.8K frames & Real & Street Camera &  \textcolor{red}{\xmark} & \textcolor{red}{\xmark} & \textcolor{green}{\cmark} \\
        \midrule
        SYNTHIA \cite{ros2016synthia} & 2016 & 13.4K frames & Synthetic & Street Camera & \textcolor{green}{\cmark} & \textcolor{green}{\cmark} & \textcolor{red}{\xmark} \\
        SynTraC \cite{chen2024syntrac} & 2024 & 6 hours & Synthetic & Street Camera & \textcolor{red}{\xmark} & \textcolor{red}{\xmark} & \textcolor{red}{\xmark} \\
        Omni-MOT \cite{sun2020simultaneous} & 2020 & 14M+ frames & Synthetic & Street Camera & \textcolor{red}{\xmark} & \textcolor{red}{\xmark} & \textcolor{red}{\xmark} \\
        SHIFT \cite{sun2022shift} & 2022 & 2.5M frames & Synthetic & On-Board & \textcolor{red}{\xmark} & \textcolor{green}{\cmark} & \textcolor{red}{\xmark} \\
        \textbf{R3ST} & \textbf{2024} & 80K+ frames & Synthetic & \textbf{Street Camera} & \textcolor{green}{\cmark} & \textcolor{green}{\cmark} & \textcolor{green}{\cmark} \\
        \bottomrule
    \end{tabular}
    }
    \vspace{0.2cm}
    \caption{Comparison of size, nature and additional annotations of existing perception datasets. R3ST is the only dataset that leverages the advantages of synthetic data generation, combining it with realistic trajectories.}
    \label{tab:trajectory_datasets}
\end{table*}

\section{The R3ST Dataset}

\begin{figure*}[!h]
    \centering
    \begin{subfigure}{0.24\textwidth}
        \includegraphics[alt={Simulated urban intersection with cars navigating through a multi-lane crossing.}, width=\linewidth]{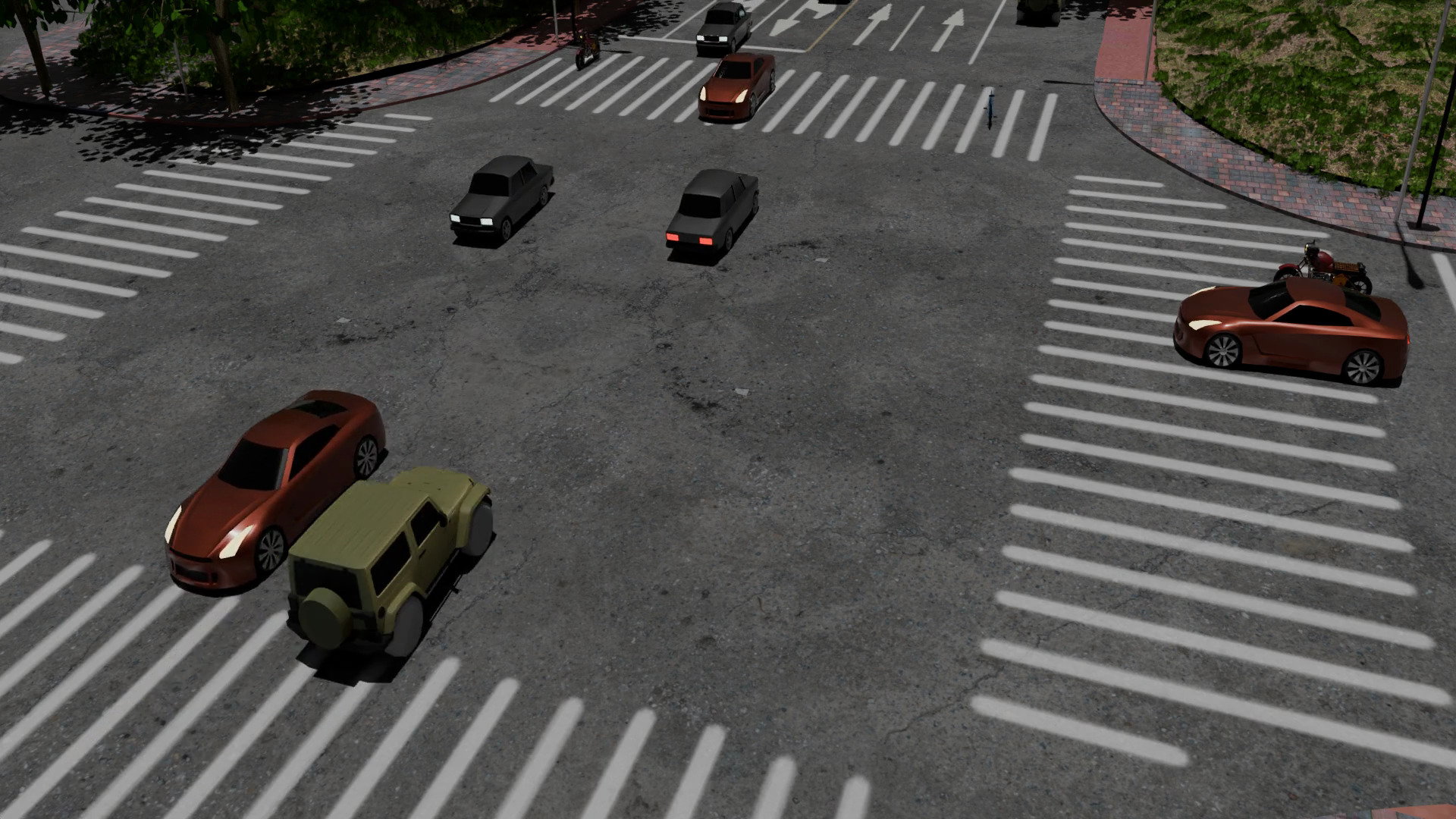}
    \end{subfigure}
    \begin{subfigure}{0.24\textwidth}
        \includegraphics[alt={Segmentation mask of simulated urban intersection with cars navigating through a multi-lane crossing.}, width=\linewidth]{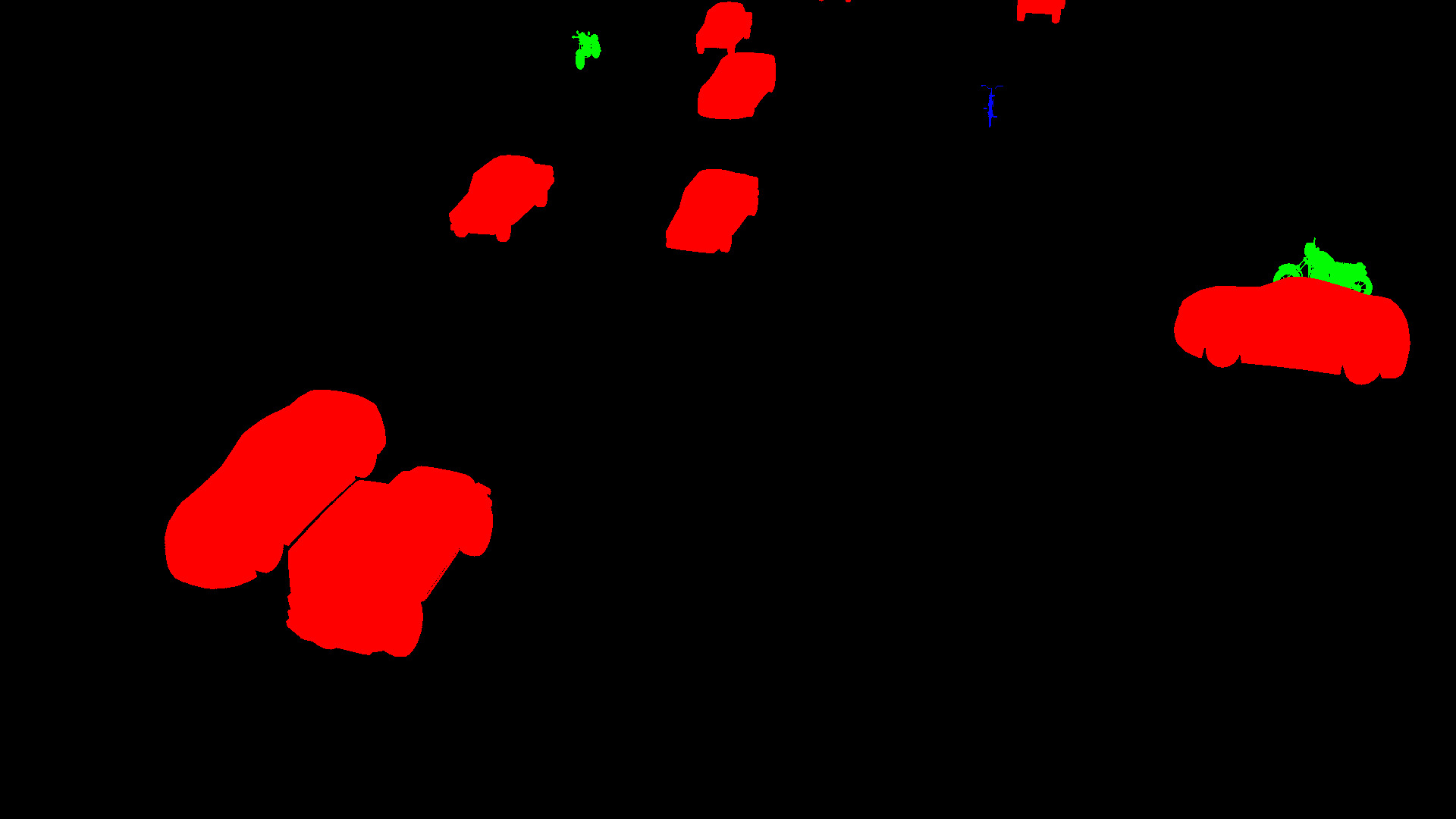}
    \end{subfigure}
    \begin{subfigure}{0.24\textwidth}
        \includegraphics[alt={Segmentation mask of urban intersection with cars navigating through a multi-lane crossing.}, width=\linewidth]
        {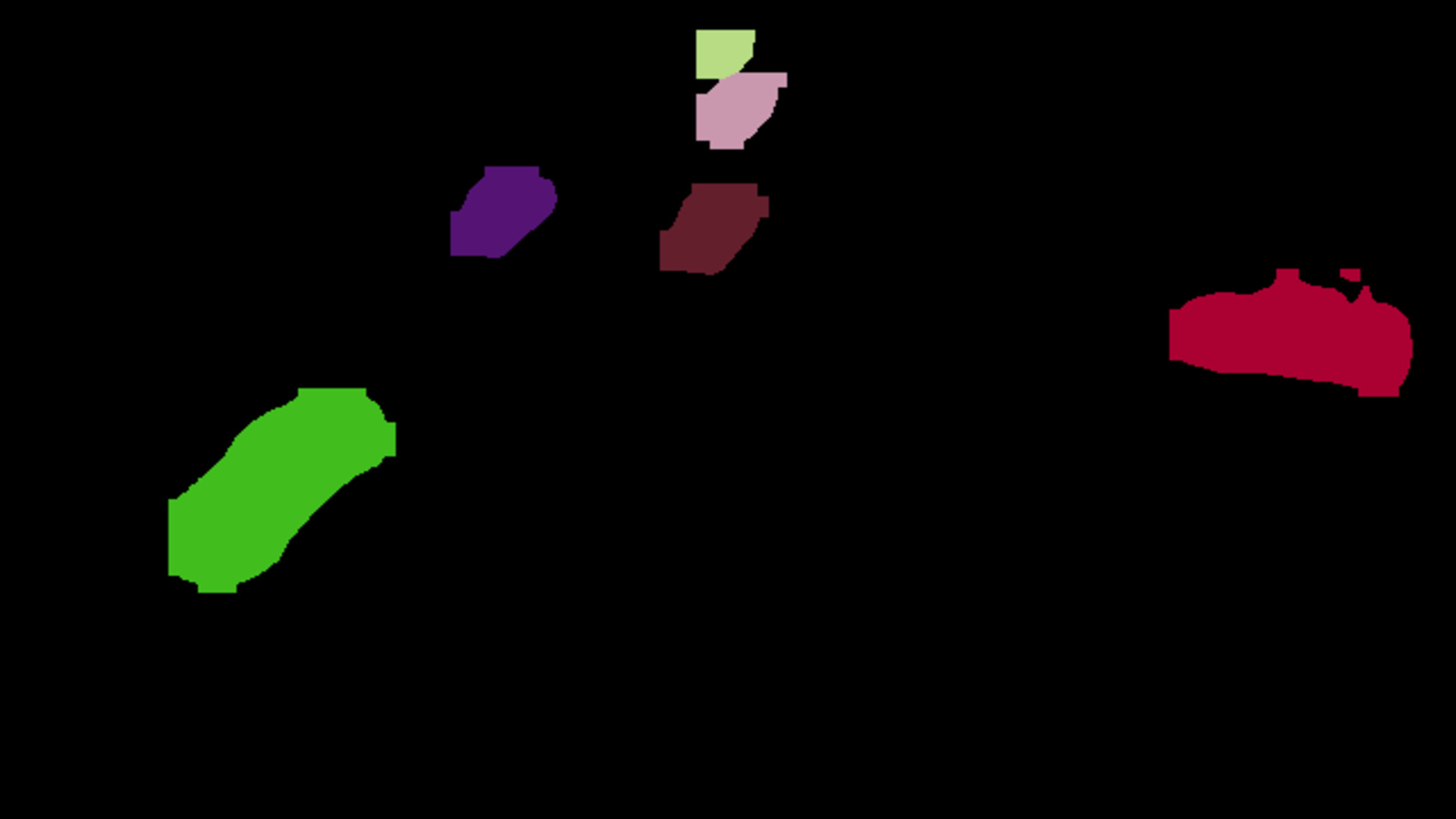}
    \end{subfigure}
    \begin{subfigure}{0.24\textwidth}
        \includegraphics[alt={Segmentation mask of urban intersection with cars navigating through a multi-lane crossing.}, width=\linewidth]
        {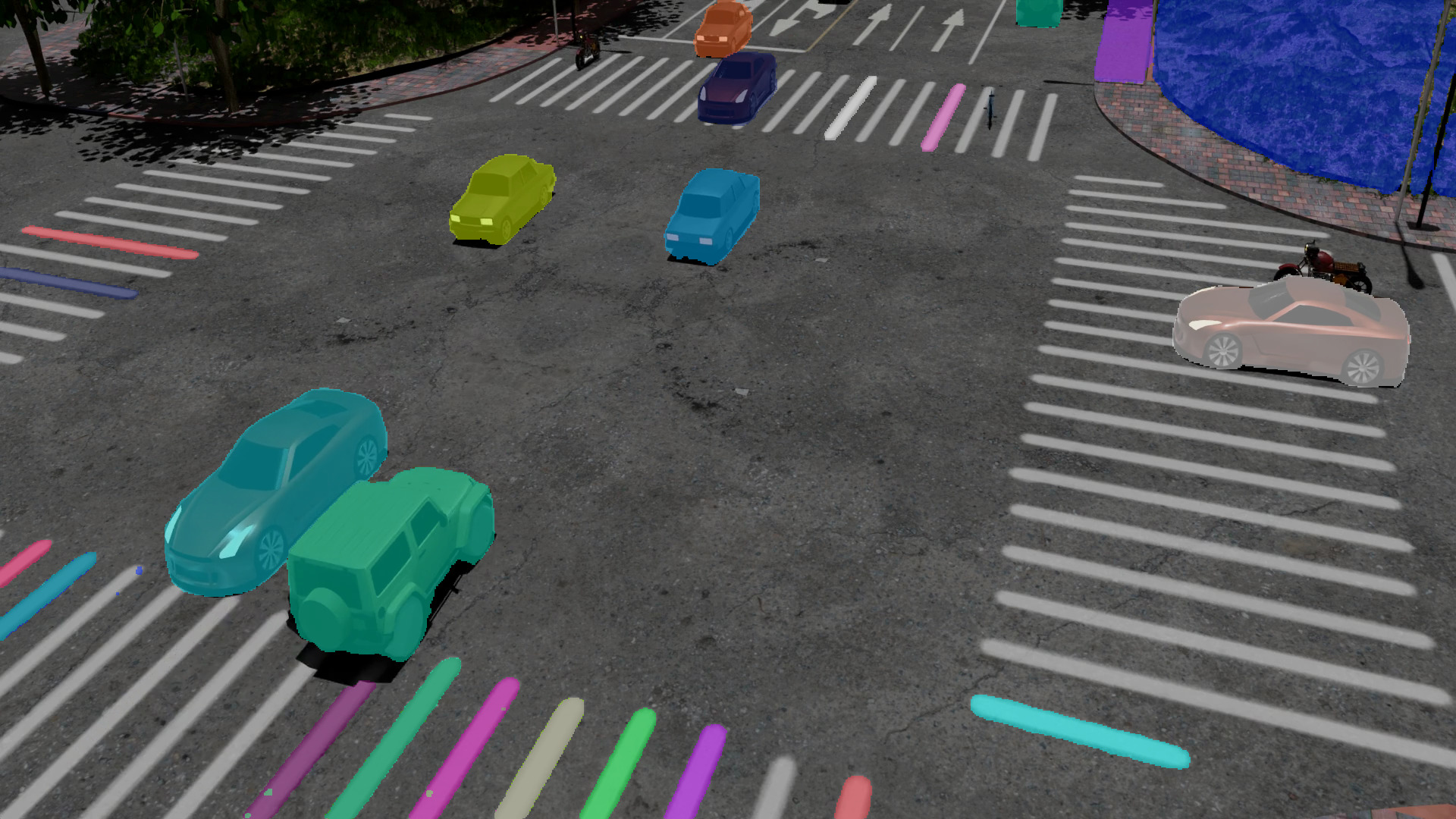}
    \end{subfigure}
    
    % \vspace{0.2cm} % Space between rows

    \begin{subfigure}{0.24\textwidth}
        \includegraphics[alt={Simulated environment urban intersection with bus navigating through a multi-lane crossing.}, width=\linewidth]{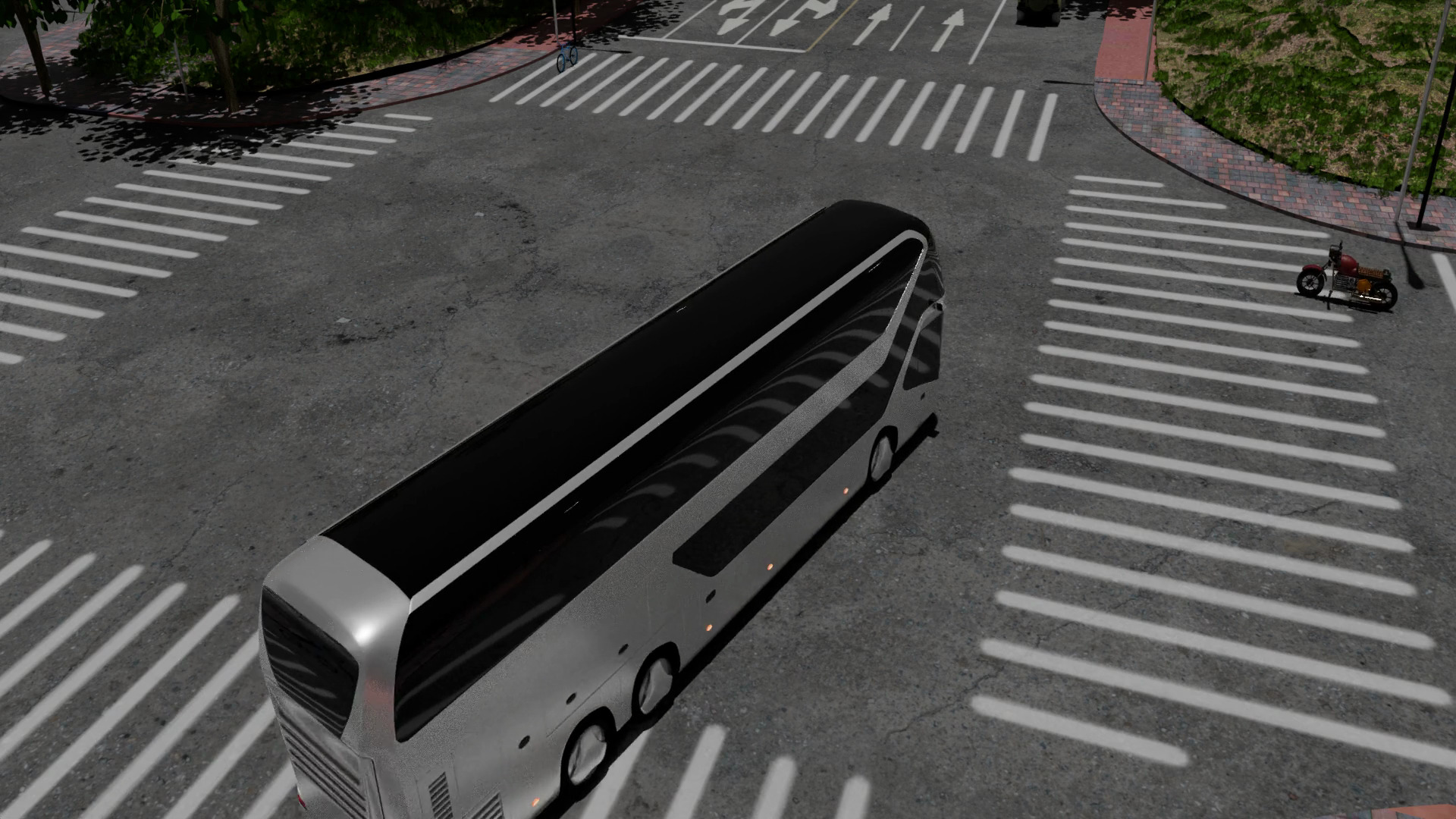}
        % \caption{RGB Frame}
    \end{subfigure}
    \begin{subfigure}{0.24\textwidth}
        \includegraphics[alt={Segmentation mask urban intersection with bus navigating through a multi-lane crossing.}, width=\linewidth]{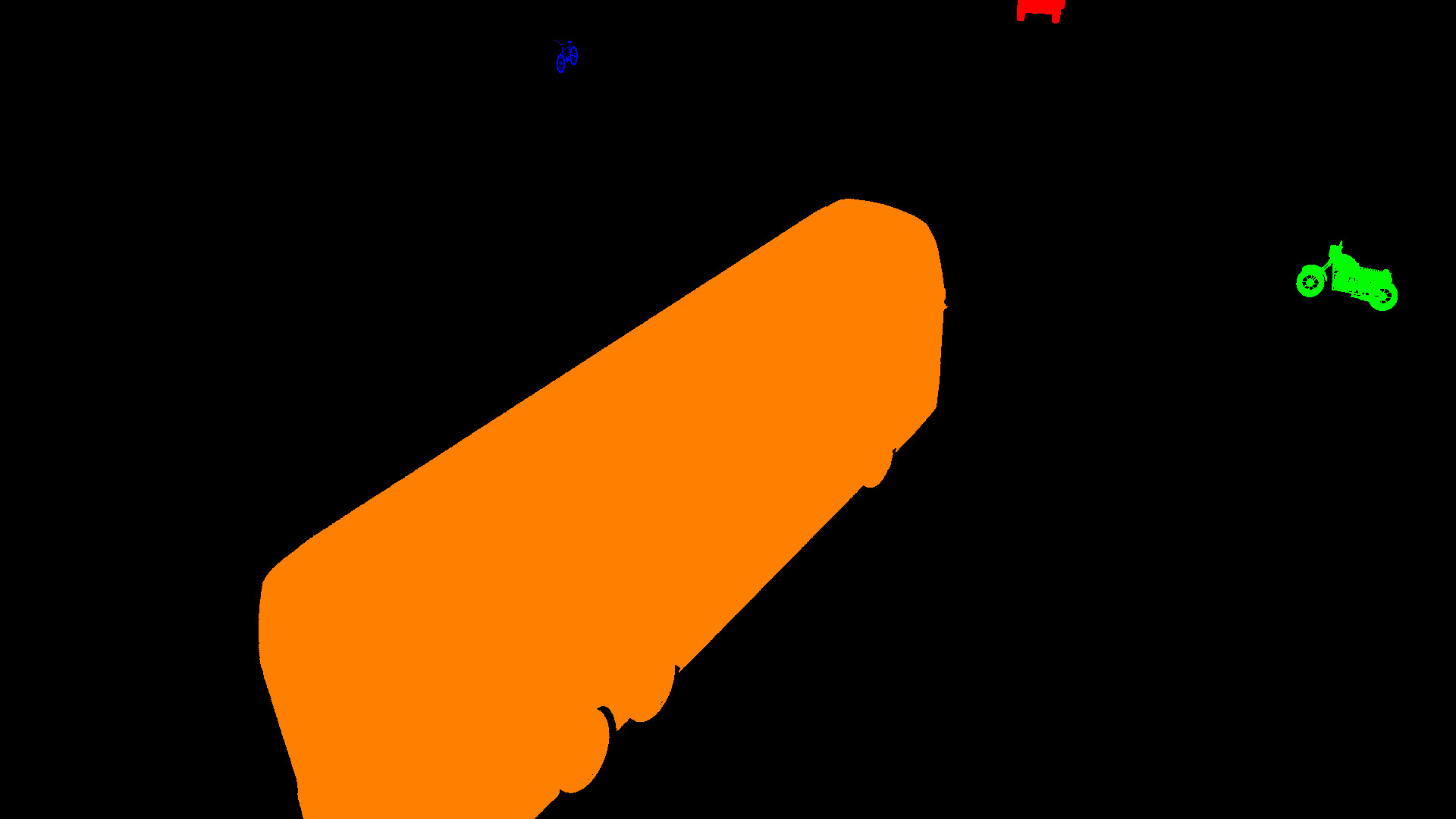}
        % \caption{Ground Truth}
    \end{subfigure}
    \begin{subfigure}{0.24\textwidth}
        \includegraphics[alt={Segmentation mask urban intersection with bus navigating through a multi-lane crossing.}, width=\linewidth]{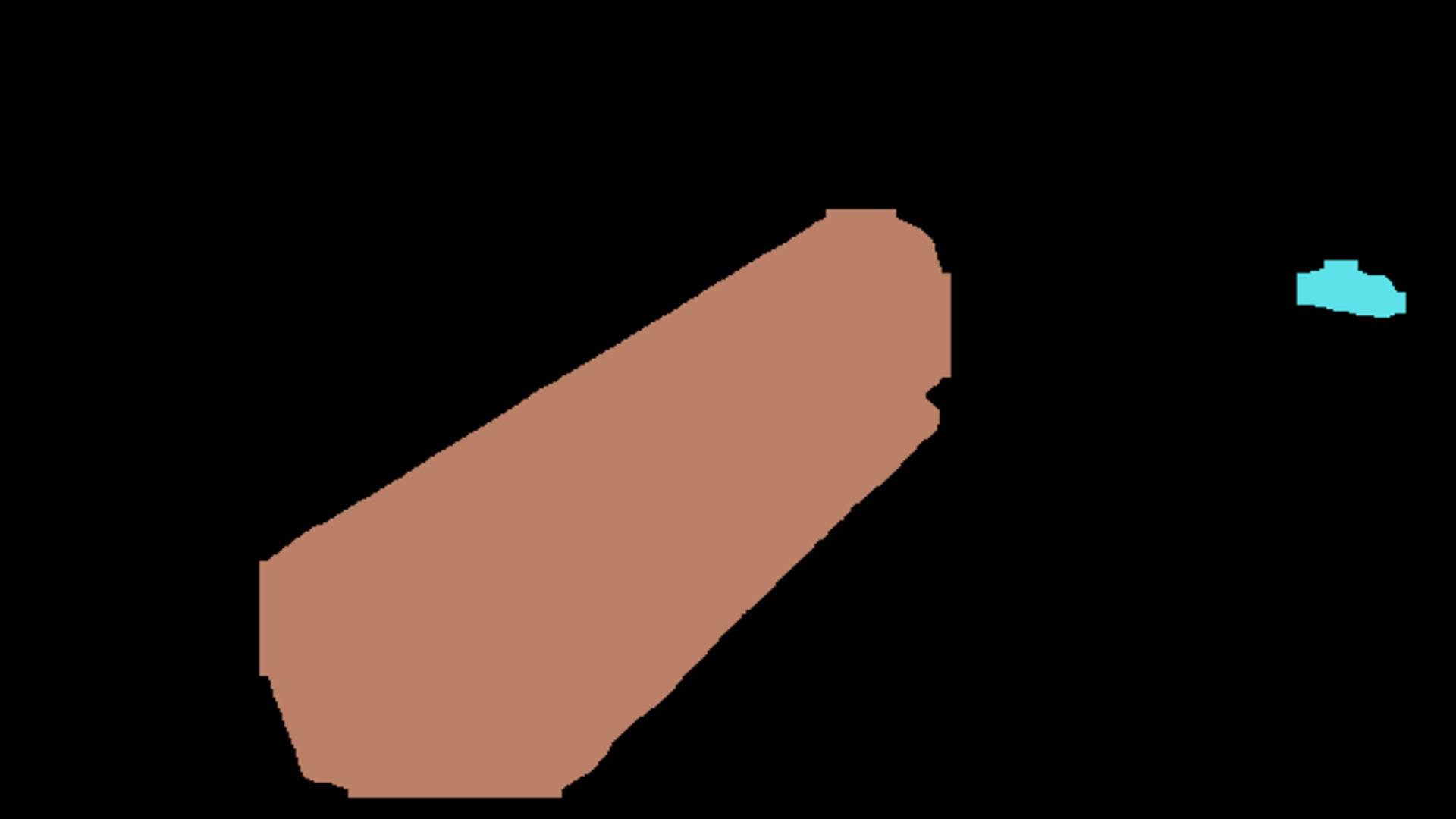}
        % \caption{YOLO11n-Seg}
    \end{subfigure}
    \begin{subfigure}{0.24\textwidth}
        \includegraphics[alt={Segmentation mask urban intersection with bus navigating through a multi-lane crossing.}, width=\linewidth]{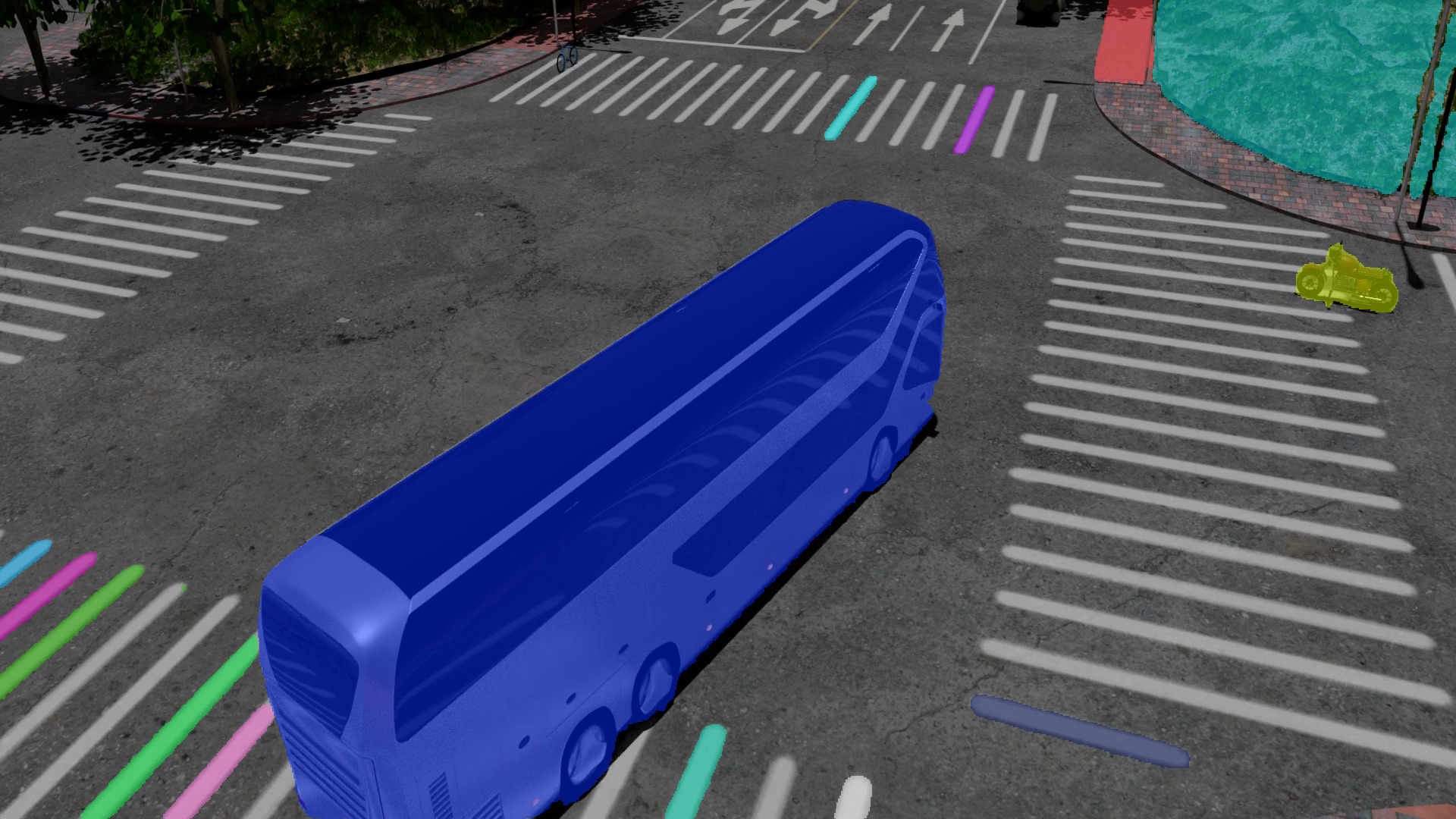}
        % \caption{SAM2}
    \end{subfigure}

    \centering
    \begin{subfigure}{0.24\textwidth}
        \includegraphics[alt={Simulated environment urban intersection with cars, bus, motorcycles navigating through a multi-lane crossing.}, width=\linewidth]{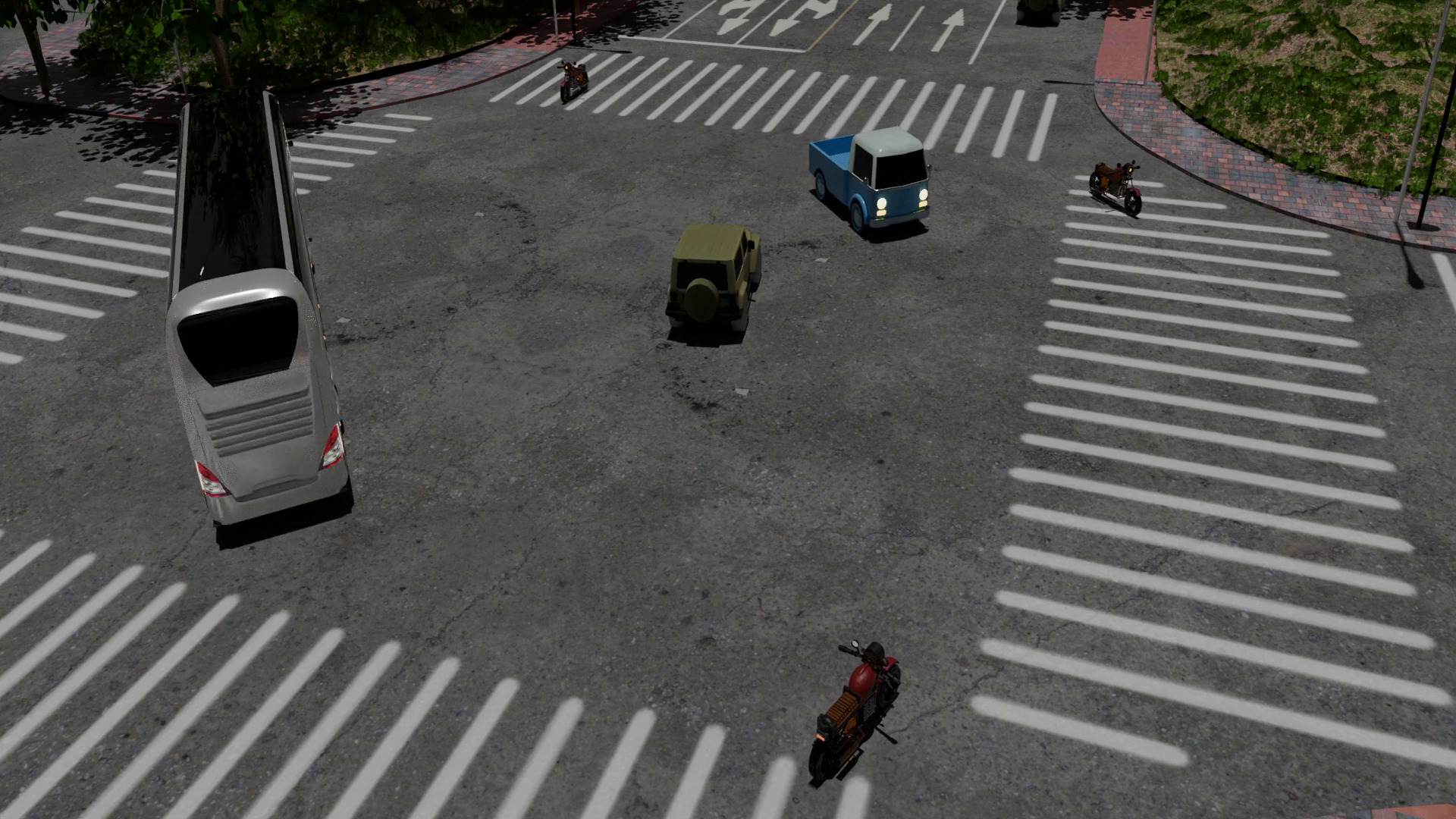}
    \end{subfigure}
    \begin{subfigure}{0.24\textwidth}
        \includegraphics[alt={Black and White Depth estimation map of urban intersection with cars, bus, motorcycles navigating through a multi-lane crossing.}, width=\linewidth]{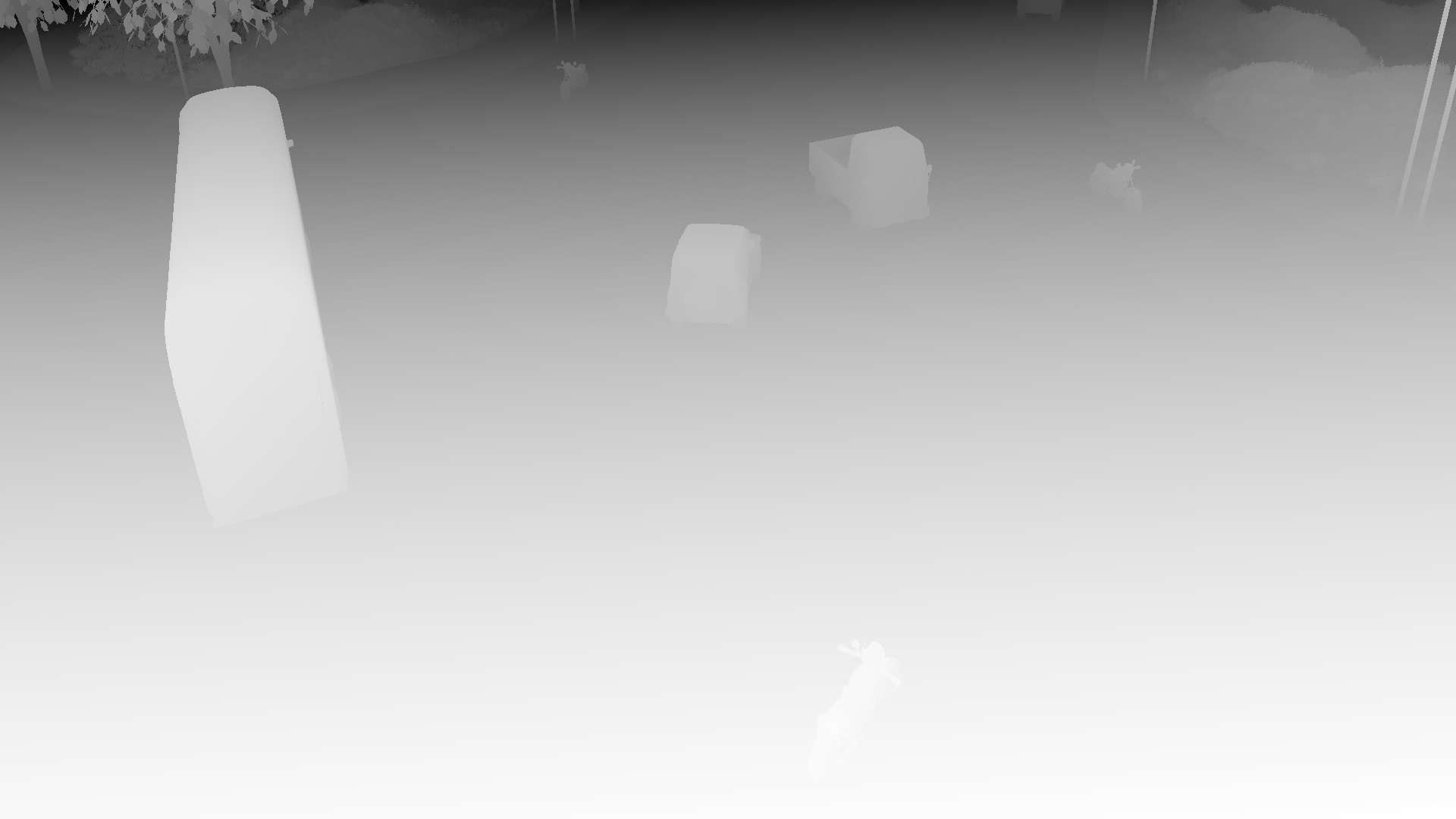}
    \end{subfigure}
    \begin{subfigure}{0.24\textwidth}
        \includegraphics[alt={Black and White Depth estimation map of urban intersection with cars, bus, motorcycles navigating through a multi-lane crossing.}, width=\linewidth]
        {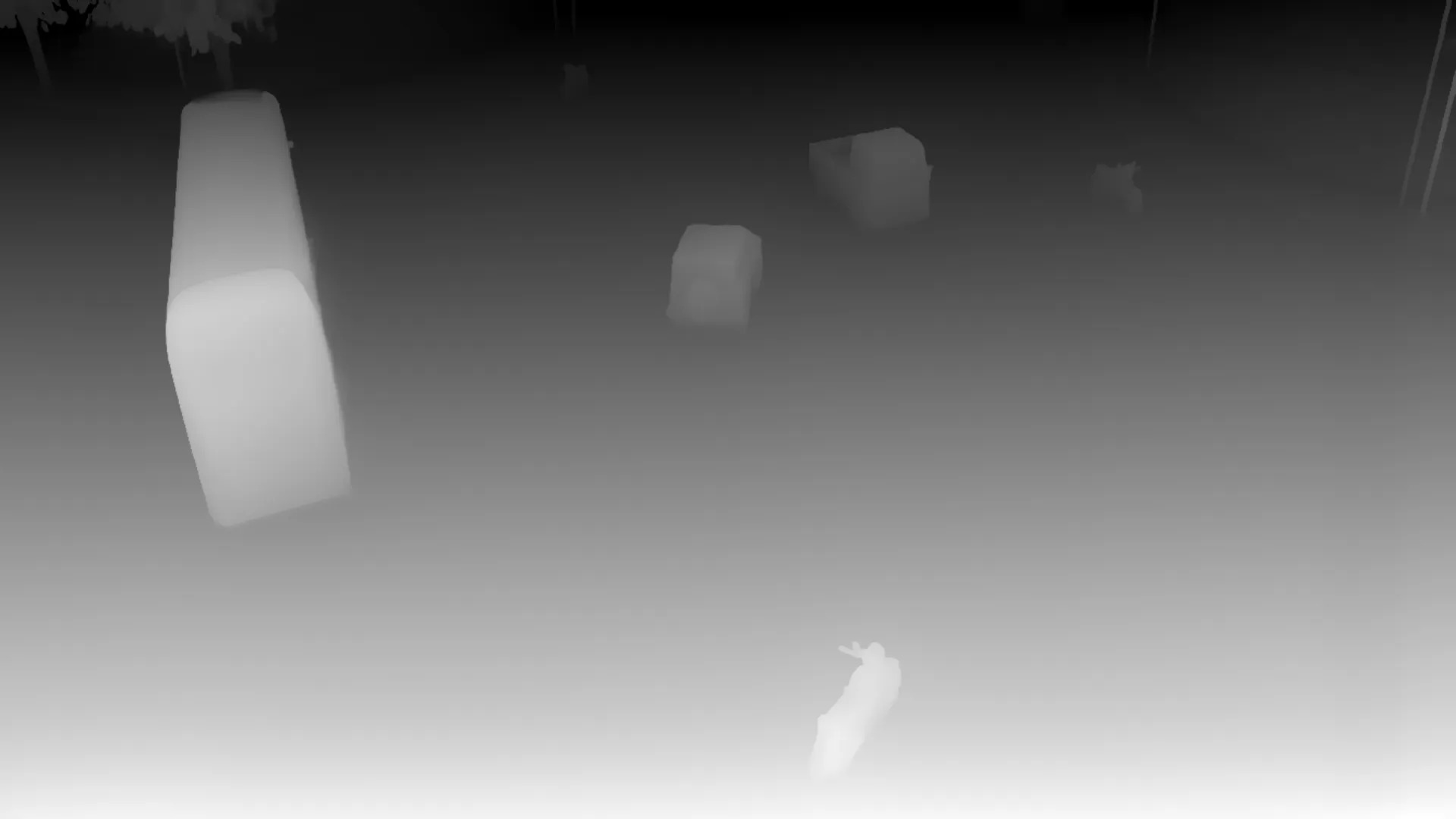}
    \end{subfigure}
    \begin{subfigure}{0.24\textwidth}
        \includegraphics[alt={Black and White Depth estimation map of urban intersection with cars, bus, motorcycles navigating through a multi-lane crossing.}, width=\linewidth]
        {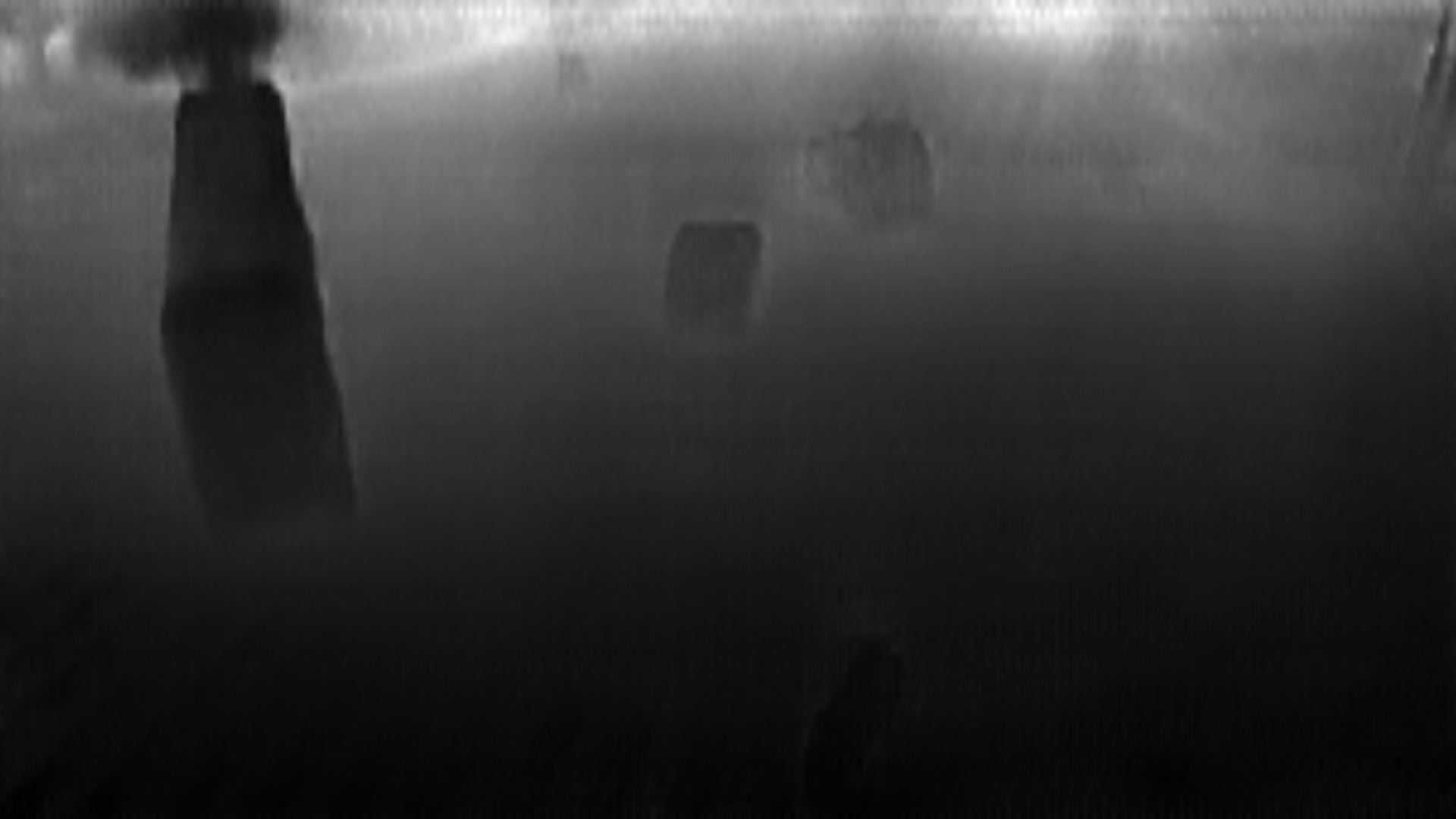}
    \end{subfigure}
    \centering
    \begin{subfigure}{0.24\textwidth}
        \includegraphics[alt={Simulated environment urban intersection with cars navigating through a multi-lane crossing.},width=\linewidth]{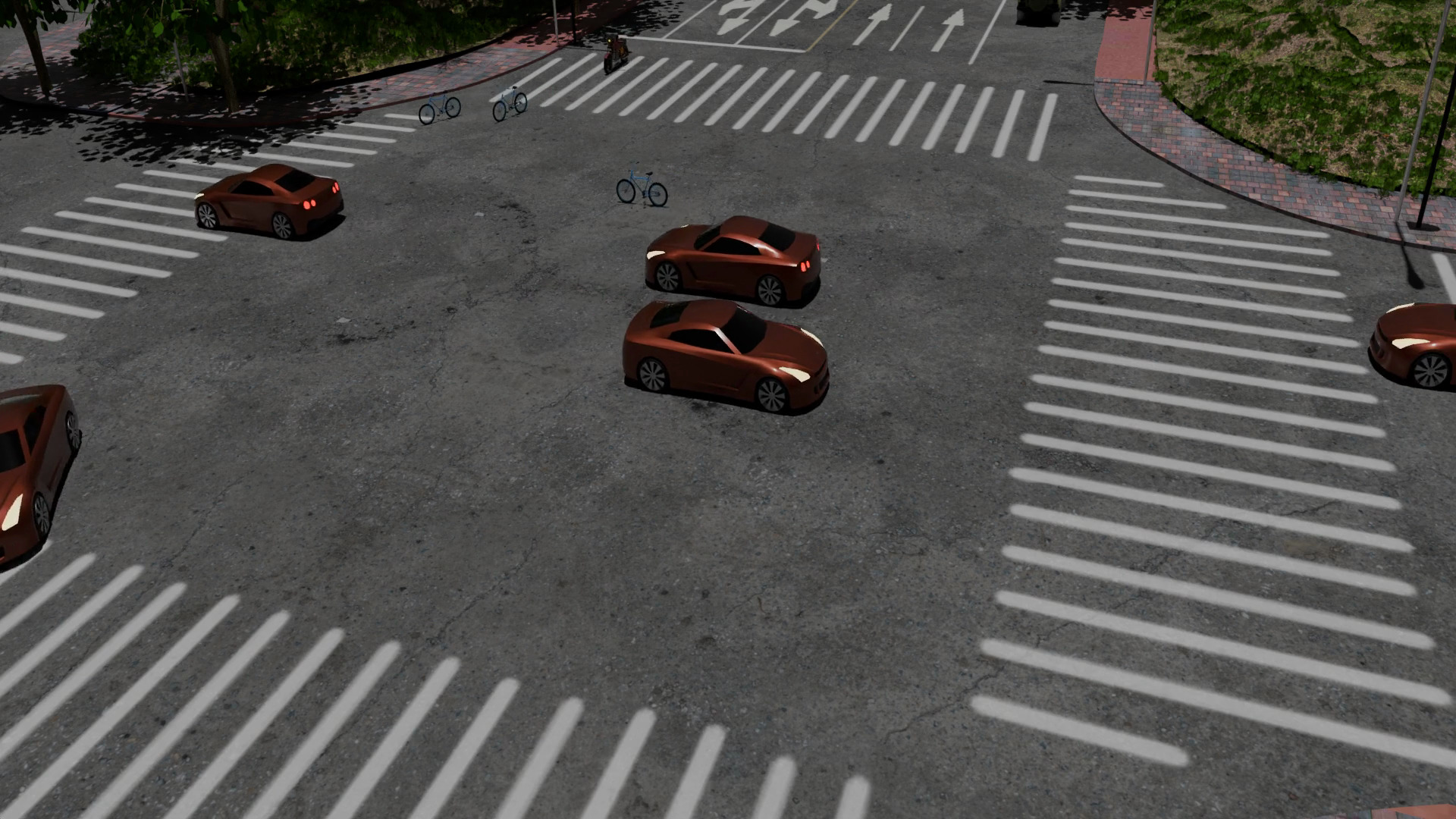}
    \end{subfigure}
    \begin{subfigure}{0.24\textwidth}
        \includegraphics[alt={Black and White Depth estimation map of urban intersection with cars navigating through a multi-lane crossing.}, width=\linewidth]{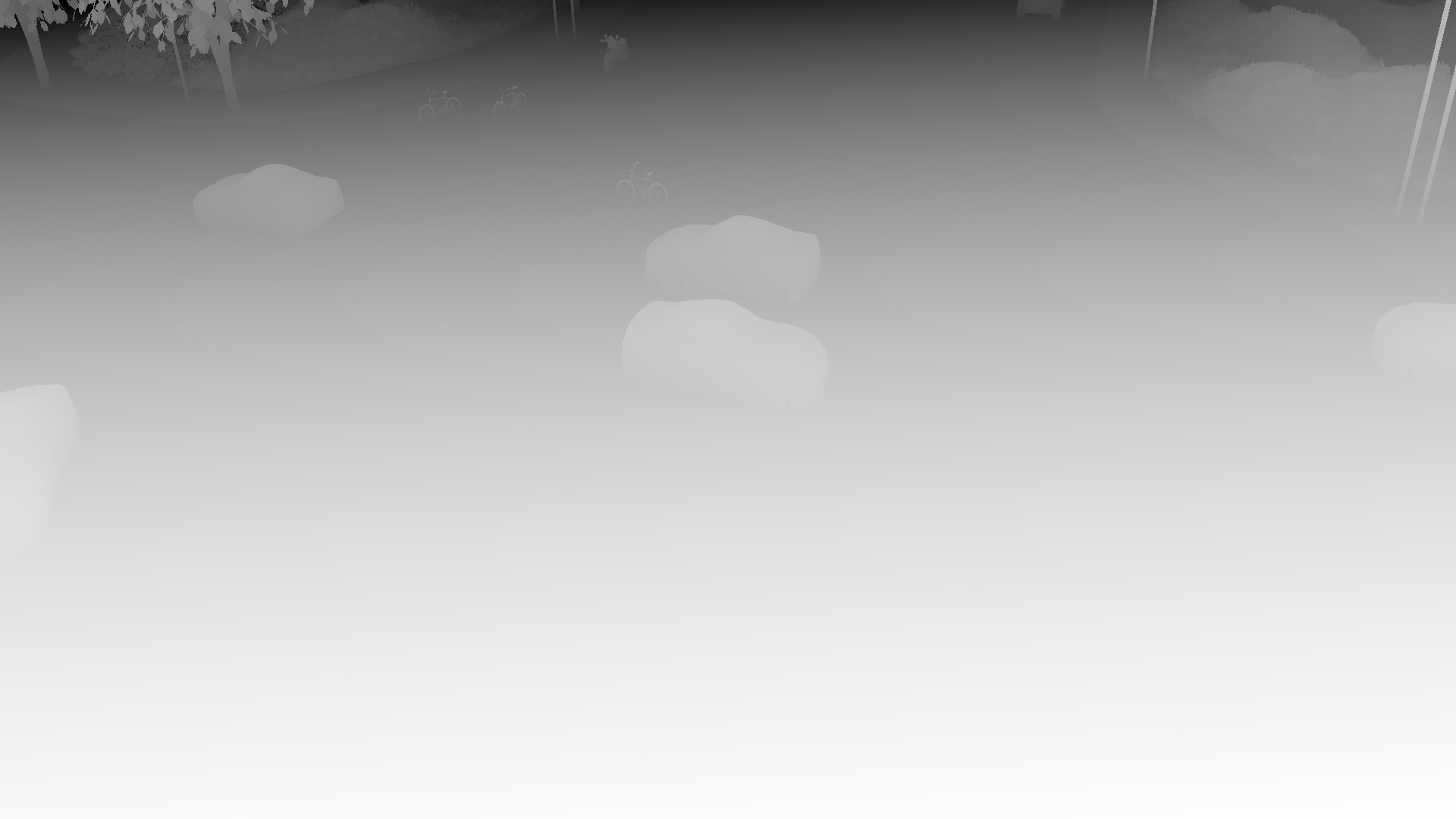}
    \end{subfigure}
    \begin{subfigure}{0.24\textwidth}
        \includegraphics[alt={Black and White Depth estimation map of urban intersection with cars navigating through a multi-lane crossing.}, width=\linewidth]{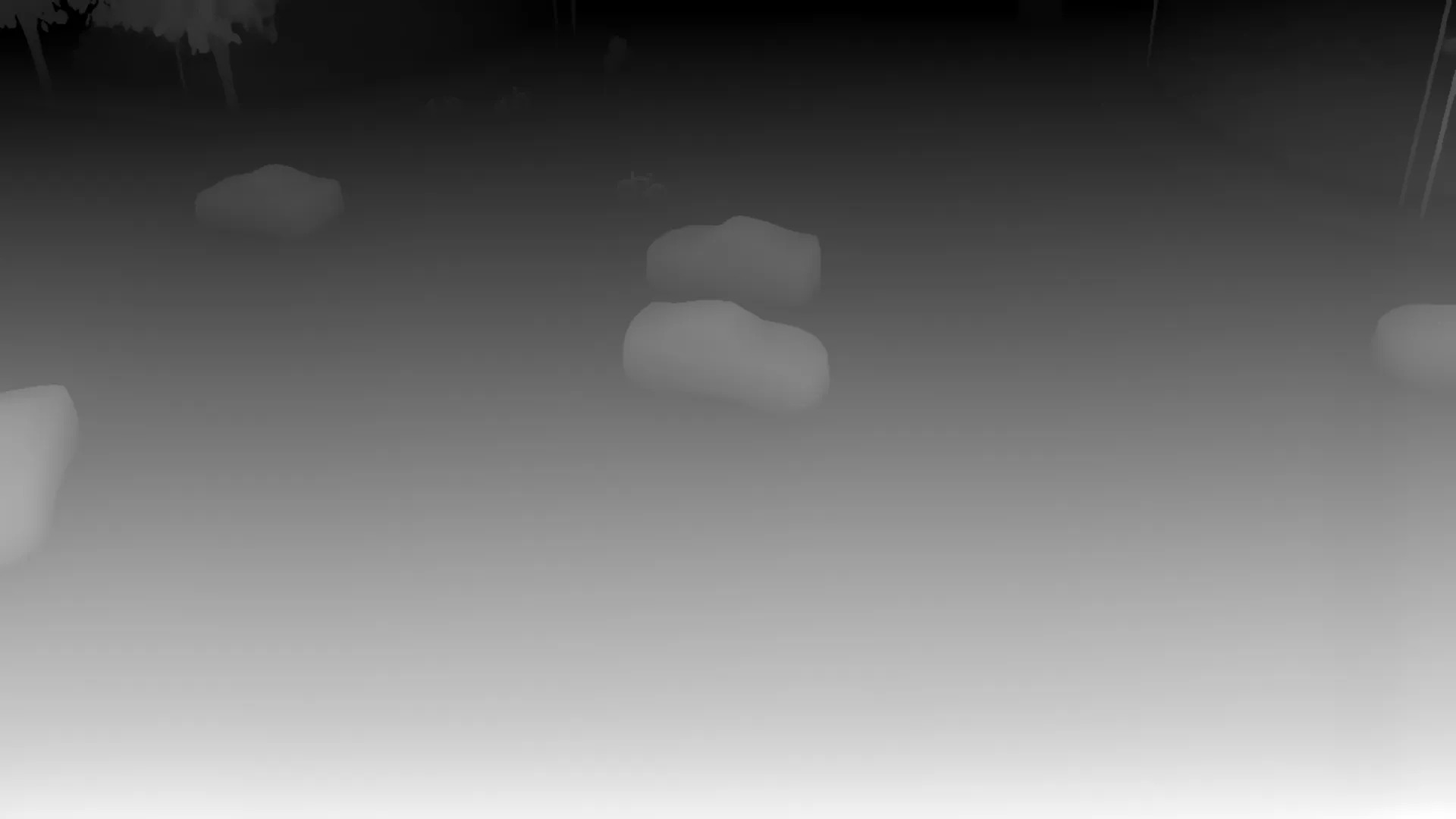}
    \end{subfigure}
    \begin{subfigure}{0.24\textwidth}
        \includegraphics[alt={Black and White Depth estimation map of urban intersection with cars navigating through a multi-lane crossing.}, width=\linewidth]
        {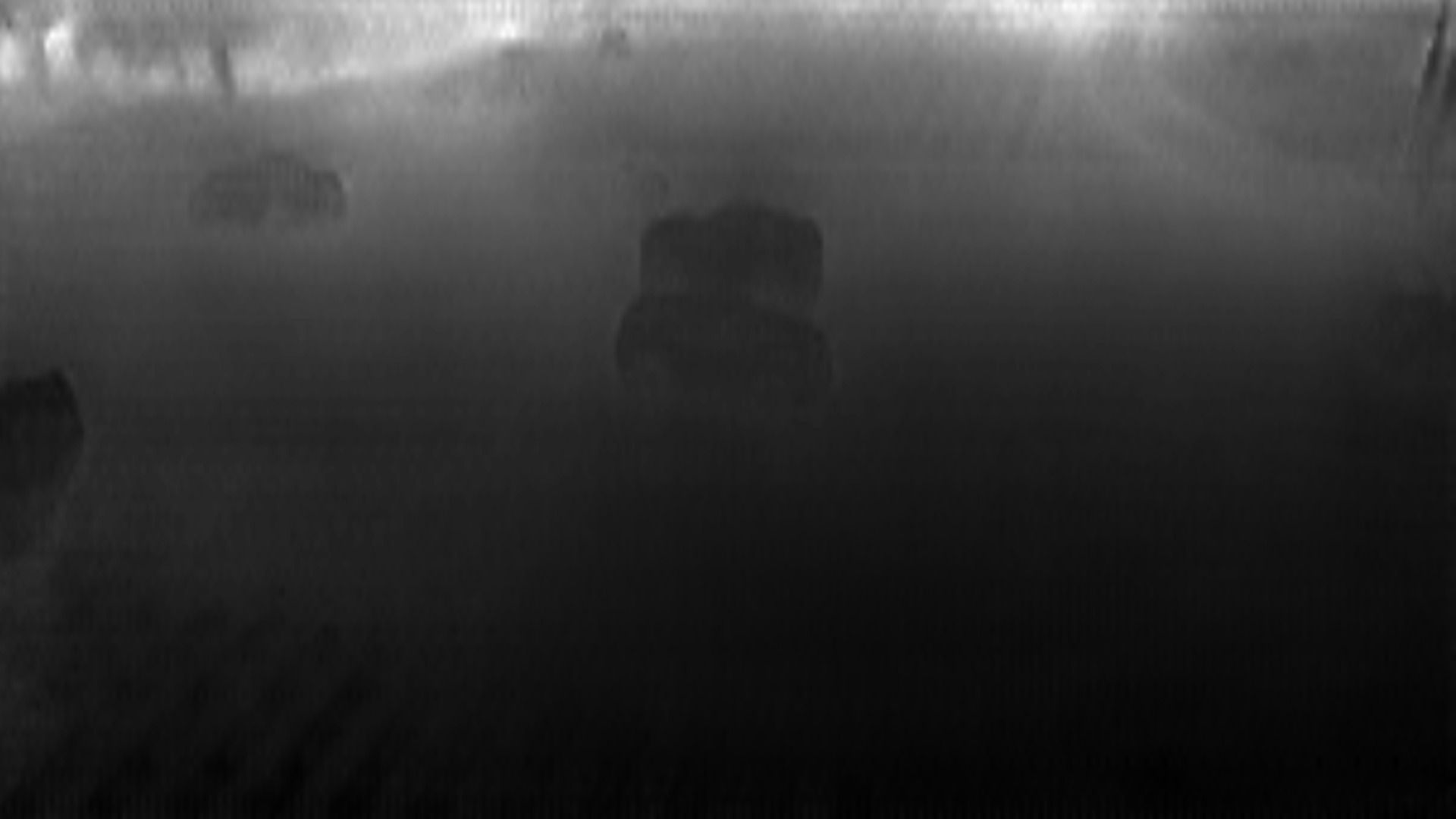}
    \end{subfigure}

    \caption{Qualitative results for instance segmentation (top two rows) and monocular depth estimation (bottom two rows). The first column shows RGB frames, the second column contains ground truth annotations, while the third and fourth columns present model predictions. Instance segmentation results are obtained using YOLO-Seg \cite{yolo11_ultralytics} and SAM2 \cite{ravi2024sam2segmentimages} online demo, while monocular depth estimation is performed with AnyDepth \cite{he2025distilldepthdistillationcreates} and Pixelformer Large \cite{agarwal2022attentionattentioneverywheremonocular} pre-trained on KITTI \cite{Geiger2013IJRR}.}

    \
    
    \label{fig:multirow}
\end{figure*}

Here we introduce the R3ST Dataset, a synthetic dataset that encompasses diverse urban intersections, multiple camera angles, and five vehicle types: cars, trucks, buses, motorcycles, and bicycles.
This dataset has been created to leverage the benefits of synthetic data while avoiding the artificial driving behaviors typical of game engines and driving simulators like Unity or CARLA. Our dataset also provides rich annotations, including instance segmentation, depth information, and bounding boxes for object detection and object tracking.
R3ST consists of photo-realistic rendering of two different intersections, each with four views, for more than 80K frames. In Table \ref{tab:trajectory_datasets} we report the properties of R3ST compared with the most famous urban scene understanding datasets, highlighting that our dataset is the only one that combines the synthetic nature of the images with real trajectories.  

\subsection{Virtual World Generation}
R3ST has been generated by rendering the virtual intersections created with Blender \cite{blender}. The motivation behind the choice of this 3D editor is the high level of freedom that it offers in creating any kind of virtual environment and the ease with which it allows the integration of external realistic trajectories. In fact, unlike typical synthetic datasets where vehicle motion is dictated by AI-driven or rule-based algorithms, R3ST incorporates real-world vehicle trajectories derived from two of the four scenarios proposed by SinD \cite{xu2022drone}, a bird’s-eye-view dataset with precise annotation of vehicle positions extracted from real drone footage. As visible in Figure \ref{fig:trajectories}, which shows the variance of clustered trajectories, this approach ensures that the motion patterns of vehicles in R3ST replicate the real-world traffic scenarios. For each trajectory of SinD, we associated the corresponding vehicle type. If the vehicle is a car we randomly selected among three different car meshes. 
To increase the realism further, we carefully selected materials for each mesh to achieve a visual fidelity comparable to real-world data. We modeled two of the four scenarios presented by SinD: an intersection in Tianjin city and an intersection in Chongqing city.
These virtual environments were created from scratch in Blender, utilizing external libraries such as BlenderKit \cite{blenderkit} to import free, realistic materials, 3D vehicle models, and building models. Additionally, we employed free add-ons like Sampling Tree Generator \cite{samplingtreegenerator}, which uses a procedural algorithm to generate vegetation. This algorithm dynamically creates trees based on user-defined parameters.
Furthermore, to simulate images captured from road cameras, we positioned them on light poles, setting the vertical sensor at 22mm, the field of view (FOV) to 35.3 degrees, and the F-Stop to 2.8. In the reconstructed scenes, four cameras were placed, each framing one of the four directions of traffic at the respective intersections. Vehicle animations were then rendered from the perspective of each camera.

\paragraph{Multimodal Annotations}
To enhance the usability of R3ST we leveraged Vision Blender \cite{cartucho2020visionblender} to compute additional multimodal annotations that can be used in a range of computer-vision applications. These per-frame annotations add information that goes beyond the scope of trajectory forecasting and are useful for different tasks like instance segmentation and monocular depth estimation (MDE). Additionally, we directly derive the 3D bounding box of each object in the scene from the Blender World Environment and project them on the image plane to get the 2D bounding boxes. We organize these annotations in YOLO format, where every object is encoded by its class label, the normalized center coordinates (cx, cy), along with the normalized width and height of the bounding box.

\section{Experiments}
We leverage multimodal annotations to demonstrate the applicability of real world tasks to our dataset. 
We aim to assess how pre-trained models perform on R3ST in tasks like Object Classification, for which we show the results in Table \ref{tab:yolo_results}, and Monocular Depth Estimation and Instance Segmentation, for which we present results in Figure \ref{fig:multirow}.

% Object detection is an important building block of TMS, allowing for accurate vehicle localization and classification. We conducted a test with a pre-trained YOLO11-large \cite{yolo11_ultralytics}, to assess its performance on our dataset. The aim is to look at how far a recent state-of-the-art object detection model generalizes to our synthetic dataset and investigate any possible domain discrepancies that could occur between synthetic and real data.
% In Table \ref{tab:yolo_results} we report the results of YOLO11-large on R3ST
% , showing great performance for a model trained on a different domain. 

% Object detection is an important building block of TMS, allowing for accurate vehicle localization and classification.
To address object detection, we tested YOLO11-large directly on R3ST, but its performance was uneven, with good precision for cars but low on motorcycles, probably due to the poor quality of the mesh used. To mitigate this, we fine-tuned YOLO11-large on R3ST for 15 epochs to show how a model trained on real images can generalize fast on our dataset. In Table \ref{tab:yolo_results} we report the resulting mAP@50 and mAP@50-95.

\begin{table}[!h]
    \centering
    \begin{tabular}{lcc}
        \toprule
        Class & mAP@50 & mAP@50-95\\
        \midrule
        Car         & 0.995 & 0.986 \\
        Van       & 0.978 & 0.925 \\
        Motorcycle  & 0.990 & 0.972 \\
        Bicycle    & 0.994 & 0.975 \\
        \midrule
        Overall     & 0.989 & 0.965\\
        \bottomrule
    \end{tabular}
    \vspace{0.2cm}
    \caption{Object detection performance of YOLO11-large on the R3ST test set, composed of 900 images.}
    \label{tab:yolo_results}
\end{table}

\section{Conclusion and Future Works}
In this paper, we introduced R3ST, the first synthetic dataset that integrates real-world vehicles trajectories, bridging the gap between realism and the flexibility of synthetic environments. We believe our dataset is valuable especially for trajectory forecasting, as well as other urban scene understanding tasks. Future works will focus on expanding the dataset with more scenes and increasing the realism of them by including different type of models for each vehicle type, as well as more challenging light conditions, to better address the domain shift problem. 

\begin{credits}
\subsubsection{\ackname}
This study has been partially supported by Italian Ministry of Enterprises and Made in Italy (Ministero delle Imprese e del Made in Italy - MIMIT) with the project PMDI under the agreements for innovation in the automotive sector D.M. 31/12/2021 and DD 10/10/2022.

Partially financed by the European Union - Next Generation EU, Mission 4 Component 1 CUP B53C23003540006. This research was partially carried out by Claudia Melis Tonti within the framework of the National Ph.D. Program funded by the Italian PNRR (Mission 4, Component 1), under Ministerial Decree No. 118/2023, in the Department of Computer, Control and Management Engineering 'Antonio Ruberti' - Sapienza University of Rome.

\end{credits}

\bibliographystyle{splncs04}
\bibliography{mybib}

\end{document}